\documentclass{article}
\usepackage{spconf,amsmath,graphicx}
\usepackage{amsmath}
\usepackage{amssymb}
\usepackage{subcaption}
\usepackage{xcolor}
\usepackage[normalem]{ulem}
\usepackage[hidelinks]{hyperref}

\definecolor{dusty_lavender}{HTML}{915C83}
\definecolor{muted_teal}{HTML}{72A98F}
\definecolor{powder_blush}{HTML}{F7A9A8}
\definecolor{yale_blue}{HTML}{0B4F6C}
\definecolor{baby_blue_ice}{HTML}{89BBFE}
\definecolor{toasted_almond}{HTML}{CA895F}

\usepackage{etoolbox}
\AtBeginEnvironment{thebibliography}{\small}

\title{Out-of-Distribution AI-generated Image Detection with Zero-Phase Deconvolution for Residual Whitening}
\name{Yusuf Yi\u{g}it Pilavc\i$^3$, Antoine Mallet$^1$, Vincent Itier$^{1,2}$, Jan Butora$^1$, Jérémie Boulanger$^1$, Patrick Bas$^1$}
\address{ $^1$Univ. Lille, CNRS, Centrale Lille, UMR 9189 CRIStAL, F-59000 Lille, France, \\
 $^2$IMT Nord Europe, Institut Mines-Télécom, Centre for Digital Systems, F-59000 Lille France, \\
 $^3$Normandie Univ, UNICAEN, ENSICAEN, CNRS, GREYC, 14000 Caen, France. \thanks{The code and data will be made available upon acceptance of the paper.}}
\begin{document}
%
\maketitle
\begin{abstract}

    Current AI-generated image detection methods struggle in out-of-distribution testing conditions because detectors are usually not robust when evaluated on generators not included in the training set.
    This work explores a way to mitigate the forensic bias between sources of images -- whether real or 
    generated -- by removing generator-specific fingerprints.
    This new training process can be approximated by pre-processing involving a deconvolution operation designed to remove correlated noise patterns inducing loss of generalization. Indeed, we show that whitening the spatial distribution
    of the image noise residuals can significantly improve the performance of detectors 
    on images coming from generators unseen during training.
    In particular, an end-to-end CNN model trained on whitened 
    image residuals improves detection by $12.6\%$ on average compared to a training involving original images.

\end{abstract}
\begin{keywords}
AI detection, generalization, noise residual, whitening, deconvolution.
\end{keywords}
%
\section{Introduction}
\label{sec:intro}

    The leaps taken by generative artificial intelligence in recent years have massively impacted
    the way people interact with multimedia content.
    In particular, the democratization of image AI-generation tools 
    has raised many serious societal concerns, such as disinformation, human manipulation, or illegal content creation~\cite{shoaib2023deepfakes}.
    In response, the ability to reliably detect AI-generated content has become a matter
    of crucial importance for the trustworthiness of the internet.
    
    More specifically, the detection of AI-generated images {via a machine learning classification} is a rather simple task when one can ensure that the
    test images originate from the same generators as the training data and when the quality of test images is important~\cite{cozzolino2024raising}.
    However, given the rapid and constant release of new models, this 
    assumption does not match practical use-cases involving test images coming from models not present in the training database.
    Therefore, much of the literature in this field is centered around 
    finding detectors having good generalization properties, which often consists in finding a representation 
    invariant to a semantic bias specific to a given generator~\cite{cozzolino2024raising, baraldi2024contrasting,guillaro2025bias},
    so that the classes are mapped to the same domain whether or not 
    they are ``in-distribution'' (ID) or ``out-of-distribution'' (OOD).

    It is important to note that the fingerprints extracted in order to accurately identify a given generator~\cite{bammey2023synthbuster,mallet2025simple}, on a counterpart may promote generalization problems on unknown generators or unknown image development pipelines since they are not aligned with the ones present in the training sets~\cite{francu2026cover}.     
    However, other biases exist, which can be termed as "forensic biases": 
    they range from color statistics~\cite{uhlenbrock2024did} to local patterns in the residuals~\cite{li2024masksim}.
    Note that, on the one hand, one can leverage these biases for detection.
    On the other hand, these biases may be source-specific (e.g., a specific generator) rather than class-specific~\cite{mallet2025simple}.
    
\begin{figure*}[t]
        \centering
        \includegraphics[width=\linewidth]{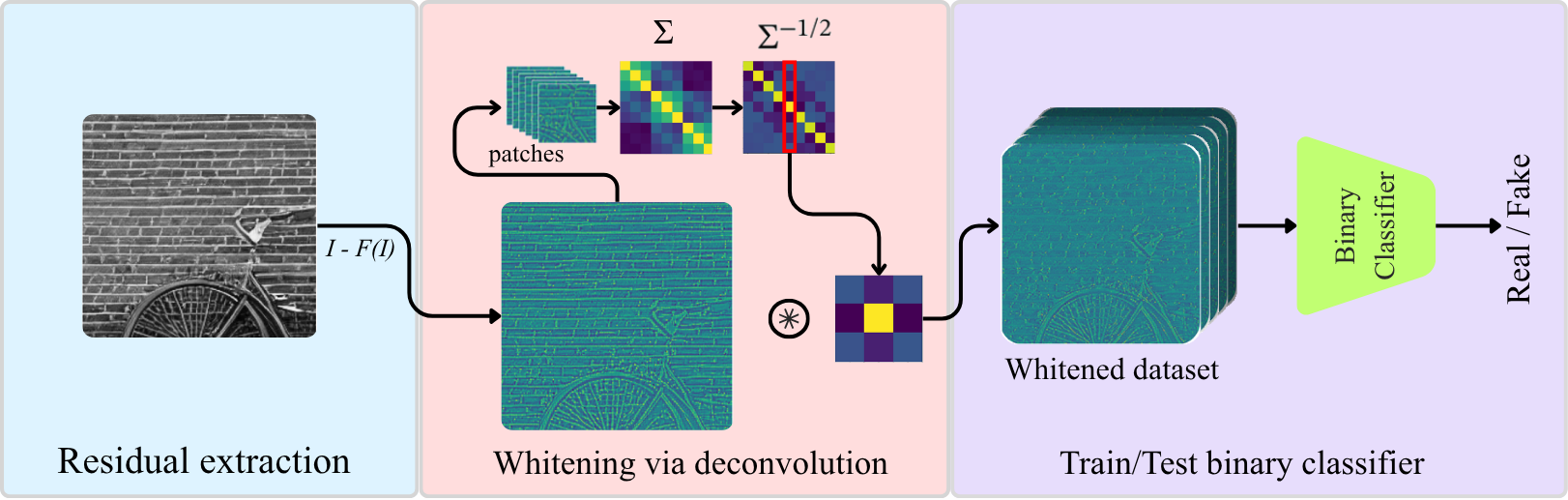}
        \caption{Overview of the proposed pipeline. Once the {noise residual $I-F(I)$} is extracted, it undergoes whitening via deconvolution proposed in~\cite{ye2019network} and the output is fed into a binary classifier during training and testing.}
        \label{fig:pipeline_overview}
        \vspace{-5mm}
    \end{figure*}

    Considering this, a safer route to generalization is to mitigate them as much as possible. In this paper, we propose a novel approach to mitigate these biases. First, we provide the forensic detector with noise residuals rather than actual images, thereby minimizing semantic bias. Furthermore, we demonstrate that removing source-specific local correlations from these residuals further reduces forensic bias and improves generalization to previously unseen AI image generators.
    This approach is motivated by established findings that discriminative patterns vary across different sources~\cite{bammey2023synthbuster, li2024masksim,mallet2025simple}. By focusing on source-specific local correlations, we present a proof-of-concept method demonstrating that aligning the local statistics of different generators enhances generalization performance. The main components of this new methodology are depicted in Fig.~\ref{fig:pipeline_overview}.

    The rest of the paper is organized as follows.
    Section~\ref{sec:sota} reviews the literature on bias mitigation for AI-generated image detection
    and positions the present paper w.r.t. existing works.
    Section~\ref{sec:whitening} then presents the proposed methodology.
    Experimental results are presented in Section~\ref{sec:results}, and Section~\ref{sec:conclusion} discusses future directions and concludes. 

\section{Related work}
\label{sec:sota}
Ideally, one would be able to detect any previously unseen generator by modeling the distribution of real images (\emph{i.e.} images acquired by a camera) and measuring a statistical deviation from this distribution. 
Unfortunately, the real distribution is impossible to model due to its diversity~\cite{li2025detecting}.
Consequently, two main approaches for OOD detection in image forensic scenarios exist. 
Firstly, a holistic framework whose goal is to find a mixture of relevant distributions (\emph{i.e.}, coming from different generators and a variety of real images) to deal with the heterogeneity of the test set~\cite{abecidan2022using}. 
Secondly, an atomistic approach assumes that the distribution of the test set can be approximated~\cite{abecidan2026tackle}. The latter approach is efficient in the steganalysis context because the image preprocessing pipeline can be approximated and applied to build a new, statistically aligned training set.
Nevertheless, image generation is costly and produces a large variety of noise fingerprints~\cite{corvi2023intriguing}. This motivates our proposal for a lightweight method to align image statistics.

\section{Residual whitening via deconvolution}
\label{sec:whitening}


Given an image $I\in\mathbb{R}^{H\times W}$,\footnote{The color dimension is omitted in this work for simplicity.} its denoised version after applying a denoising filter $F$ is denoted as $F(I)\in\mathbb{R}^{H\times W}$. The resulting image noise residual is then expressed as $R = I - F(I)$.
Here, $F$ can be chosen as any denoising filter from the vast literature of denoisers~\cite{elad2023image}.

\subsection{Local Covariance}
We unfold the residual $R$ into overlapping $k\times k$ patches with $k \ll H, W$ and compute the covariance $\Sigma$ of these patches following the method described in~\cite{ye2019network}.
We reorganize $R$ into patches via $\texttt{im2col}$ operator:
\begin{equation}
\begin{split}
    \texttt{im2col}(R)[i,:] &= \big[R[r_i : r_i+k-1,\ c_i : c_i+k-1]\big] \in \mathbb{R}^{k^2}, \\ \qquad  i &= 1, \dots, (H-k+1)(W-k+1),    
\end{split}
\end{equation}
where $(r_i, c_i)$ is the top-left corner of the $i$-th patch.
Each patch is vectorized into $k^2$-dimensional vectors and placed at the corresponding row of $\texttt{im2col}(R)$. 
\begin{figure}[t!]
    \centering
    \begin{subfigure}[b]{0.6\linewidth}
        \centering
        \includegraphics[width=\linewidth]{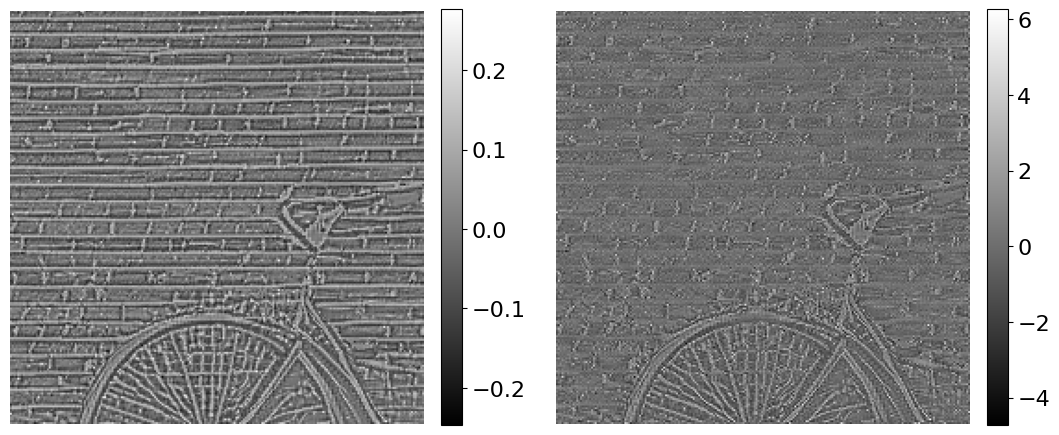}
        \caption{}
        \label{fig:whitening-before-after}
    \end{subfigure}
    \hfill
    \begin{subfigure}[b]{0.35\linewidth}
        \centering
        \includegraphics[width=\linewidth]{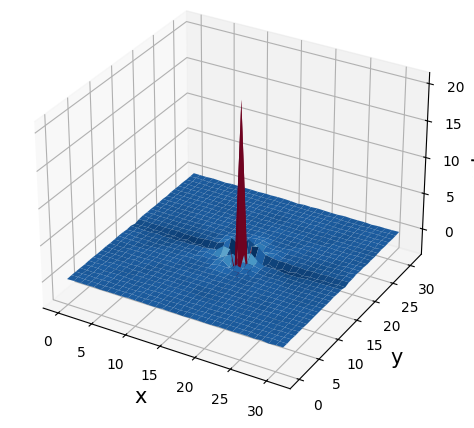}
        \caption{}
        \label{fig:whitening-kernel}
    \end{subfigure}
    \begin{subfigure}[b]{\linewidth}
        \centering
        \includegraphics[width=\linewidth]{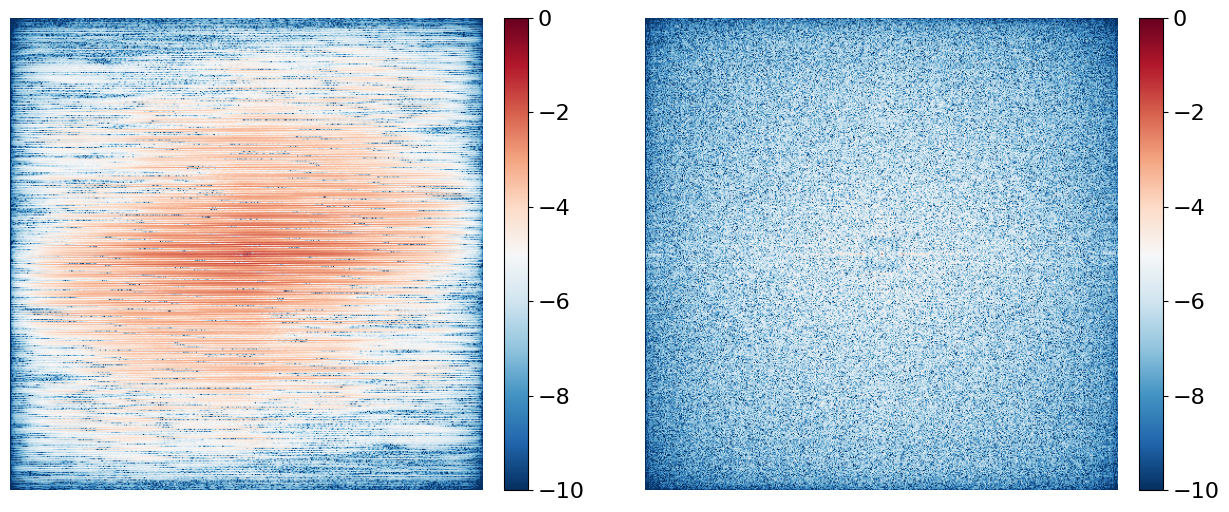}
        \caption{}
        \label{fig:whitening-autocorr}
    \end{subfigure}
    \caption{An illustration of deconvolution via inverse covariance. (a) Residual extracted from an image before and after whitening, (b) the $32\times 32$ convolutional filter in 3D, (c) the residual's autocorrelation (in $\log$ magnitude) before and after whitening.}
    \label{fig:whitening-conv-pipeline}
    \vspace{-5mm}
\end{figure}
\begin{figure*}[t]
\centering
\begin{subfigure}[b]{0.32\textwidth}
    \centering
    \hspace{-1cm}
    \includegraphics[width=1.05\linewidth]{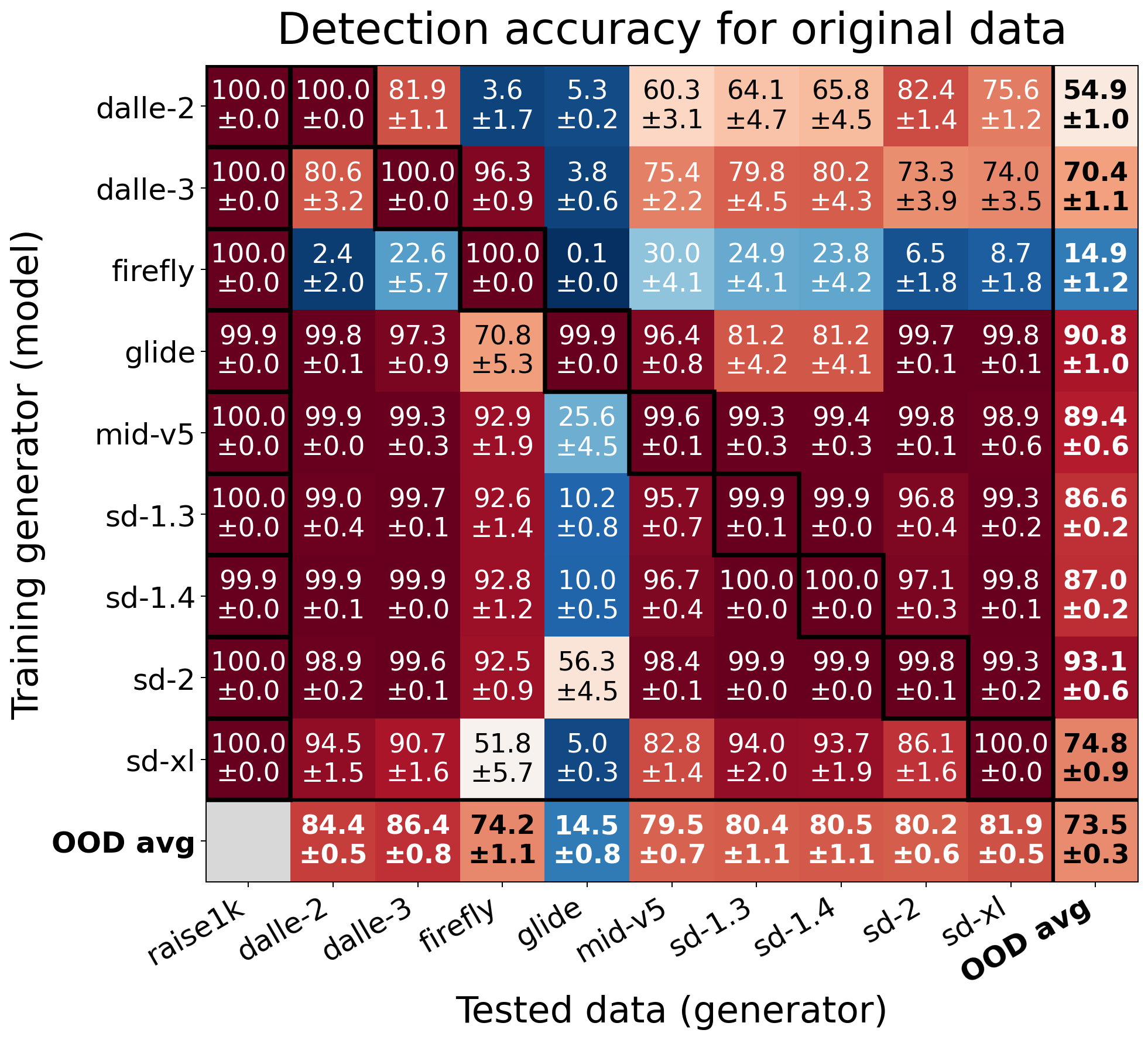}
    \caption{}
    \label{fig:acc-original}
\end{subfigure}
\hfill
\begin{subfigure}[b]{0.32\textwidth}
    \centering
    \hspace{-1cm}
    \includegraphics[width=1.05\linewidth]{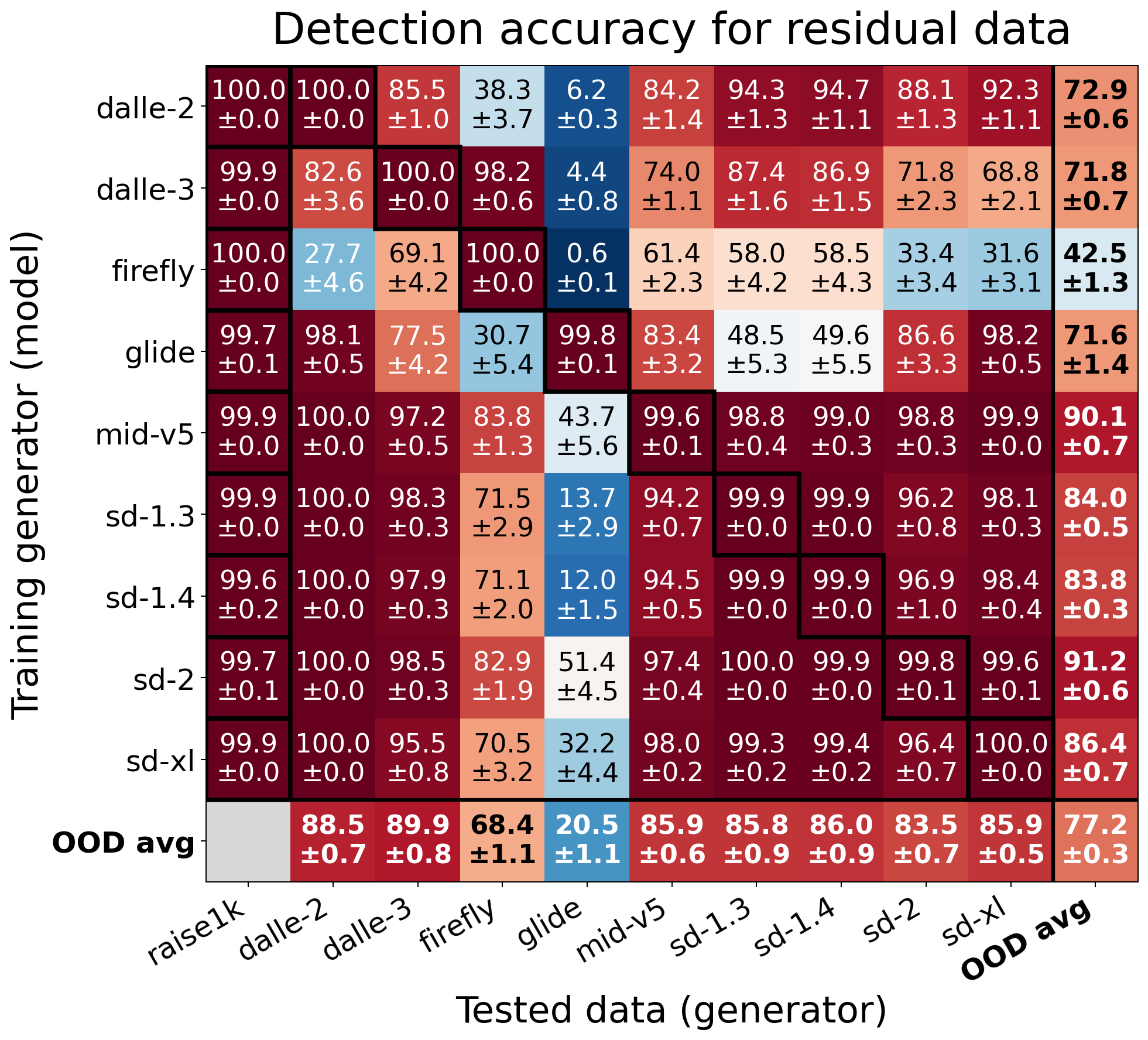}
    \caption{}
    \label{fig:acc-residual}
\end{subfigure}
\hfill
\begin{subfigure}[b]{0.32\textwidth}
    \centering
    \hspace{-1cm}
    \includegraphics[width=1.05\linewidth]{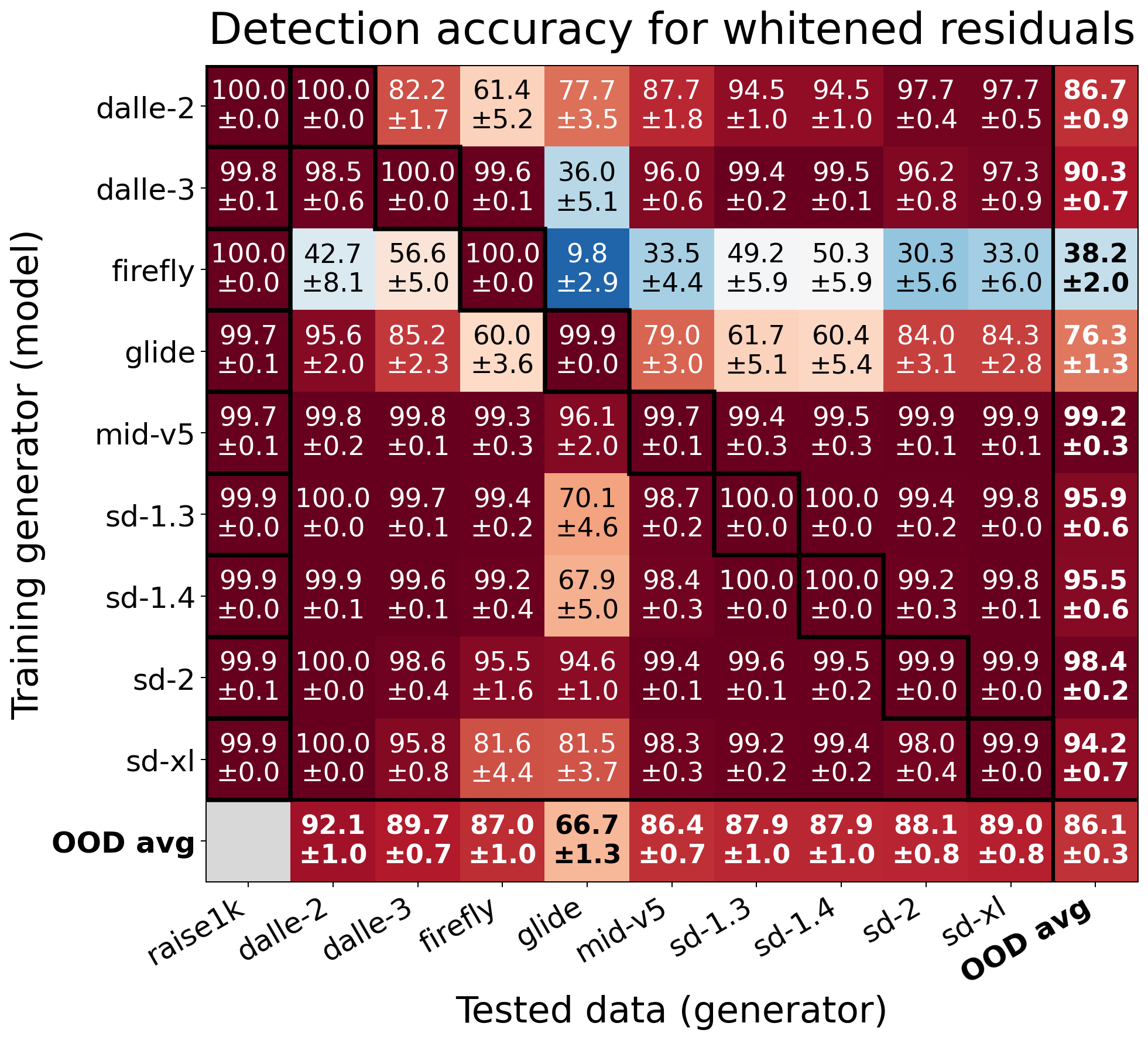}
    \caption{}
    \label{fig:acc-whitened}
\end{subfigure}
\caption{Classification accuracy of DJIN trained on the original images (a), residuals (b), and whitened residuals (c). The in-distribution accuracies are highlighted in bold black frames. The results in off-diagonals (after the first column) correspond to OOD accuracies.
Each table reports the average OOD accuracy.
}
\label{fig:acc-heatmaps}
\vspace{-5mm}
\end{figure*}

The covariance matrix of this representation is the sample covariance matrix of all $N$ patches of $R$: 
\begin{equation}
   \Sigma = \frac{1}{N-1}{\texttt{im2col}(R)^\top \texttt{im2col}(R)} \in \mathbb{R}^{k^2 \times k^2}. 
\end{equation}
Note that $\texttt{im2col}(R)$ is centered before computing $\Sigma$.
\subsection{Local Whitening}

The zero-phase shift whitening~\cite{kessy2018optimal} of $\texttt{im2col}(R)$ is
\begin{equation}
    S= \texttt{im2col}(R)\Sigma^{-1/2}. 
    \label{eq:applyinverse}
\end{equation}
This type of whitening avoids possible rotation or sign ambiguities, which is the case for PCA-like whitening~\cite{huang2018decorrelated}.
Once the whitened representation $S$ is obtained, it is not evident how to reproduce the original image representation in the size of $H\times W$. 
The authors in~\cite{ye2019network} propose an approximation by taking the central column of $S$ as the whitened output. 
Then, Eq.~\eqref{eq:applyinverse} shrinks to $\texttt{im2col}(R)\texttt{c}(\Sigma^{-1/2})$ where \texttt{c} slices the central column of $\Sigma^{-1/2}$. 
This product can be seen and calculated as a convolution between $R$ and a filter $w$ which is obtained by reshaping $\texttt{c}(\Sigma^{-1/2})\in\mathbb{R}^{k^2}$ into a $k\times k$ filter.\footnote{The calculation by the convolution may be advantageous when $\Sigma$ is estimated from randomly sampled patches, which is not exploited in this work.}  
Finally, the whitened image $\tilde{R}$ is produced by convolution $R\circledast w$.
In the rest, we call $w$ the deconvolutional filter.

Fig.~\ref{fig:whitening-conv-pipeline} illustrates the residual whitening described above. 
Once the deconvolution kernel is computed and applied to $R$, we also compute the autocorrelation function $A(\Delta x, \Delta y)$ of the image before and after whitening via a 2D Fourier transform.  
The deconvolutional filter removes most correlations, mainly between adjacent pixels, without introducing new long-distance ones. 
In Fig.~\ref{fig:whitening-kernel}, the overall shape of the kernel is shown to be similar to a Laplacian filter, hence removing positive correlations.
Moreover, we observe strong horizontal directionality in the filter, consistent with the mainly horizontal correlations in the given example.

\section{Experimental results}
\label{sec:results}

\noindent \textbf{Procedure and metrics.} In the experimental section, we focus on the use of whitened residuals for the classification task: \textit{Real} or \textit{Fake}?
First, we take a convolutional deep AI detector, DJIN~\cite{doi2025deepfake}, which uses the SRNet backbone pre-trained on millions of images for steganalysis tasks~\cite{JIN21}. 
We train three separate instances of DJIN: on natural images, and unwhitened and whitened residuals. 
Each classifier is trained for 10 epochs on a pair of real and fake sources (e.g., \texttt{raise1k} and \texttt{dalle-3}), then tested on the seen (in-distribution) and unseen generators (ODD).
We averaged the results over $N=20$ repetitions to reduce the randomness of different classifier-weight initializations and train/validation/test splits. The final results in Fig.~\ref{fig:acc-heatmaps} are presented as mean ($\mu$) $\pm$ standard error ($\sigma/\sqrt{N}$). 
All models are trained and tested using PyTorch on a workstation with two NVIDIA RTX A6000 GPUs.
\smallskip

\noindent \textbf{Datasets.} Synthbuster~\cite{bammey2023synthbuster} contains the real images collected from the RAISE (\texttt{raise1k}) database~\cite{dang2015raise} and fake images generated via nine different diffusion models; {DALL-E 2} (\texttt{dalle-2}), {DALL-E 3} (\texttt{dalle-3}), {Adobe Firefly} (\texttt{firefly}), {Midjourney v5} (\texttt{mid-v5}), {Stable Diffusion 1.3} (\texttt{sd-1.3}), {Stable Diffusion 1.4} (\texttt{sd-1.4}), {Stable Diffusion 2} (\texttt{sd-2}), {Stable Diffusion XL} (\texttt{sd-x}l), and {Glide} (\texttt{glide}). 
For each real and fake image source, we have 997 images.~\footnote{Three images were dropped from all sources due to file corruption (one image) and numerical errors in whitening due to singular covariance matrices (two images).} 
For residual extraction, we use a GPU-compatible implementation of the Wiener filter proposed in~\cite{bled2023pushing} for the filter $F$.\footnote{https://github.com/MrBled/WienerPytorch}
Then, each residual passes through a deconvolution process, as illustrated in Fig.~\ref {fig:whitening-conv-pipeline}. 
We set the size of the deconvolutional kernel at $32\times32$.

For all classifiers, the training dataset consists of 600 real and 600 fake images/residuals. For the validation set, we keep 100 images from each class to avoid overfitting, which leaves 297 images per class for in-distribution testing.  
\smallskip

\noindent \textbf{Results.} Fig.~\ref{fig:acc-heatmaps} summarizes the main results of the paper. 
Training on unwhitened residuals improves some classifiers (e.g., the one trained with \texttt{firefly}) while degrading others (e.g., the one trained with \texttt{glide}). Overall, it leads to a limited improvement of $3.7\%$ in average OOD classification. In contrast, residual whitening improves average OOD accuracy by $12.6\%$ and $8.9\%$ compared with training on original pixels and non-whitened residuals, respectively.
It improves the performance of most classifiers, notably \texttt{dalle-2} and \texttt{dalle-3}, by $31.8\%$ and $19.9\%$ in average OOD accuracy, respectively, when compared with the classifier trained with original pixels, and by $13.8\%$ and $18.5\%$ when trained on unwhitened residuals, respectively. 
\smallskip

\begin{figure}
    \centering
    \includegraphics[width=\linewidth]{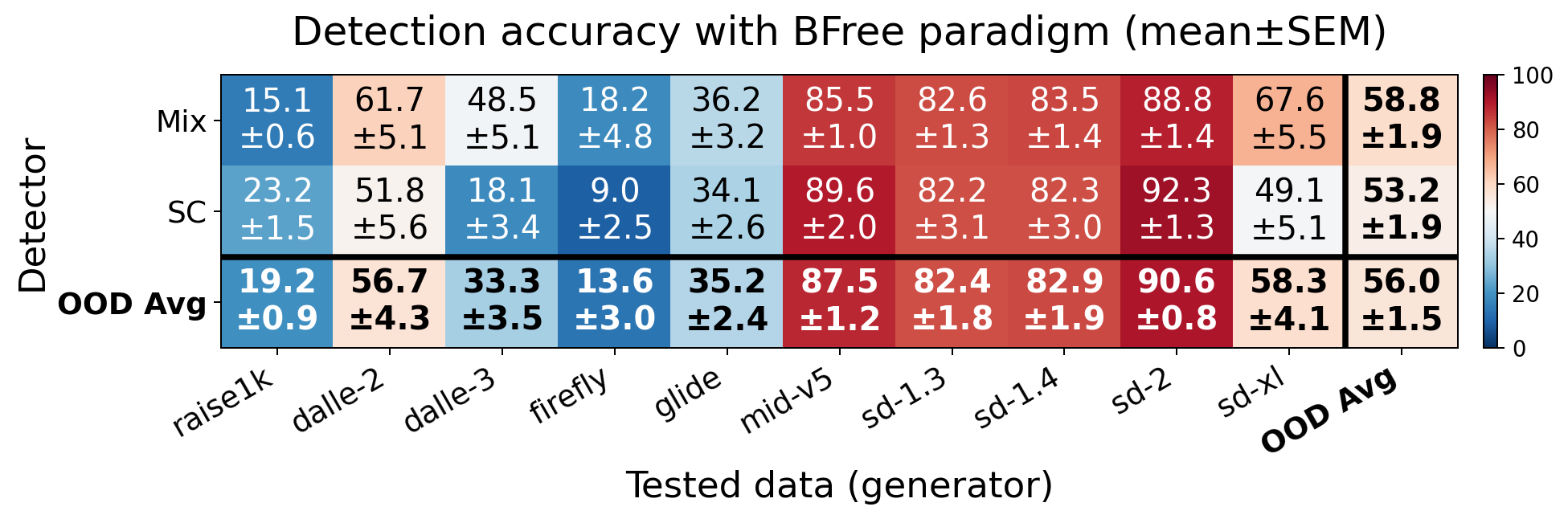}
    \caption{OOD accuracy for DJIN trained using BFree training methodology.}
    \label{fig:sota_comparison}
    \vspace{-5mm}
\end{figure}

\noindent \textbf{Comparison with SOTA.} We compare our preprocessing method to a state-of-the-art BFree (Bias-Free)~\cite{guillaro2025bias} preprocessing of training data. 
BFree paradigm generates a bias-free training dataset by generating 6 generated versions of real images from COCO dataset. This generation is based on Stable Diffusion v2.1 (SD2.1) and involves: self-conditioning, inpainting with same object category, inpainting with different category; all three coupled with original and regenerated background.  To compare with our methodology, we sample two sets, both containing 997 real images and their generated counterparts to train DJIN on. In the first set, only self-conditioned (SC) images with original background are used. In the second set, we sample a balanced mixture (mix) of all 6 generated versions.
We apply the same splitting to training, validation, and testing and measure the OOD accuracy on the Synthbuster dataset. 
The results, averaged over 20 repetitions, are presented in Fig.~\ref{fig:sota_comparison}. 
This inferior performance likely stems from the significant drop in the size of the BFree training dataset, which makes a fair comparison with our method difficult. On the other hand, our whitening preprocessing can improve generalization performance even with a small training dataset by $30\%$ compared to BFree in  OOD accuracy. 
\smallskip
\begin{figure}
\centering
    \begin{subfigure}[b]{0.47\linewidth}
        \includegraphics[width=\linewidth]{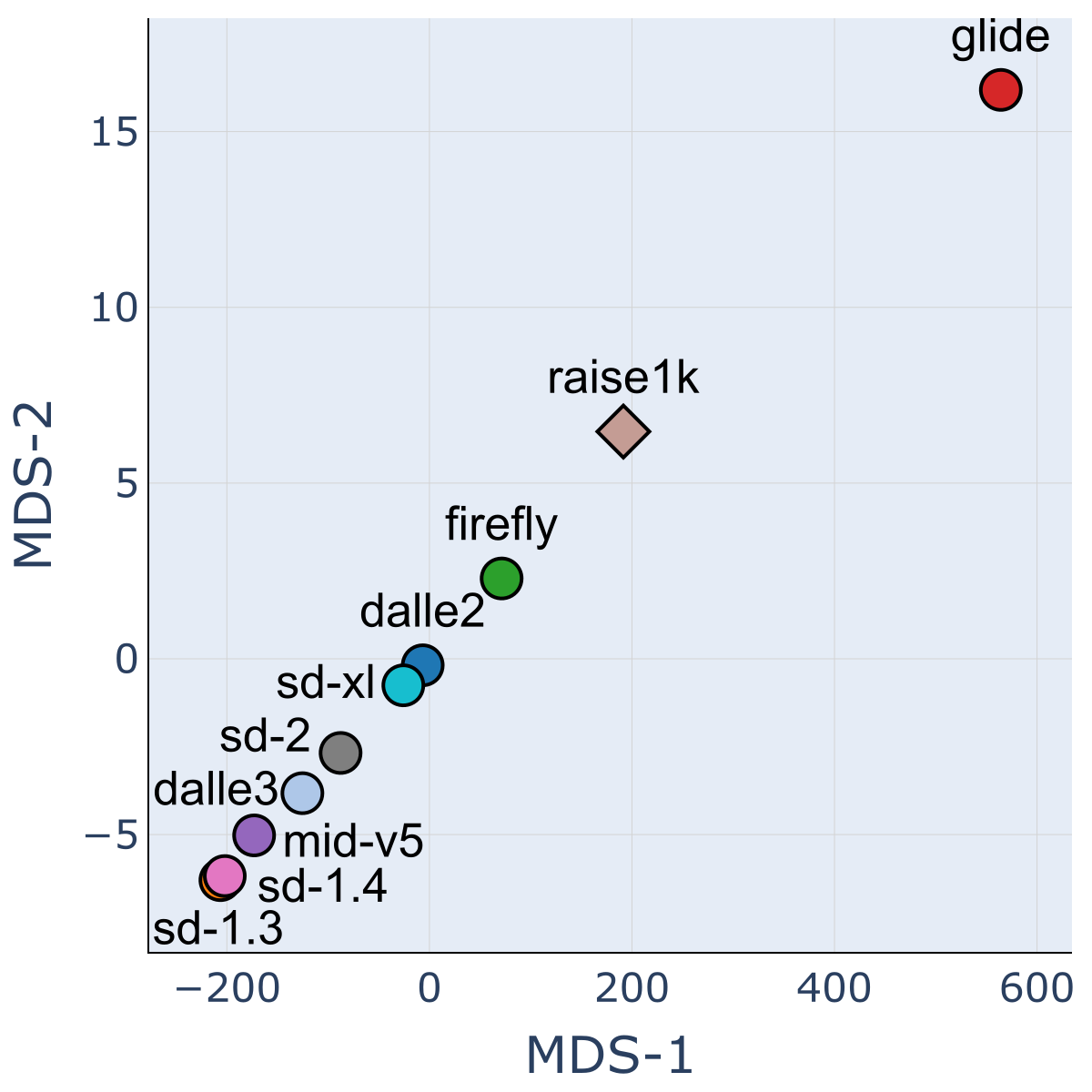}
        \caption{}
        \label{fig:mds-raw}        
    \end{subfigure}
    \begin{subfigure}[b]{0.51\linewidth}
        \includegraphics[width=\linewidth]{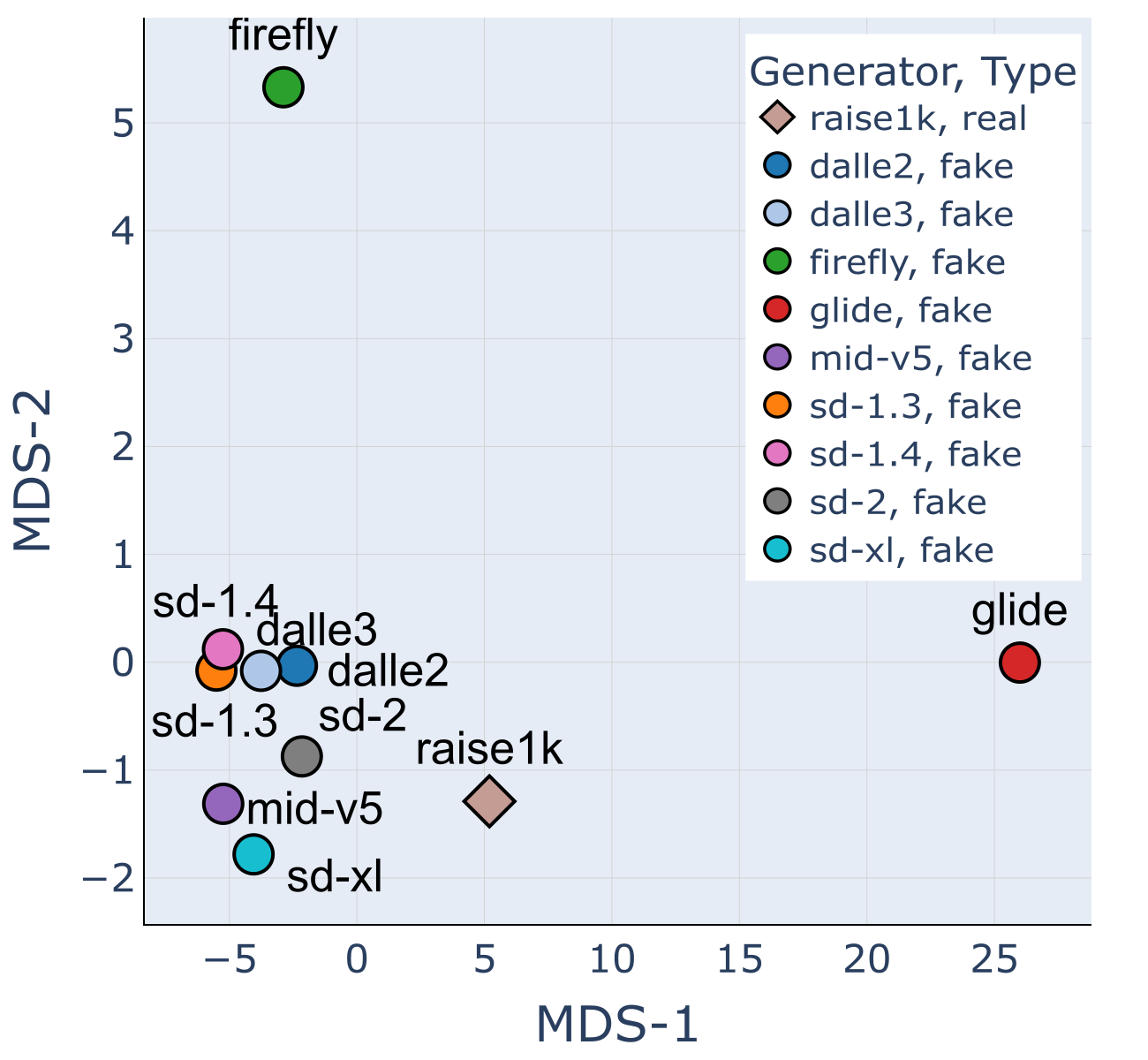}
        \caption{}
        \label{fig:mds-white}        
    \end{subfigure}
    \caption{2D MDS embeddings of the Jeffreys divergence matrix computed across generators, without whitening (a) and with whitening (b).}
    \label{fig:mds-embeddings}
    \vspace{-5mm}
\end{figure}

\noindent \textbf{Analysis of the effect of whitening.} 
For a deeper understanding of the poor performance on \texttt{firefly} and \texttt{glide}, we conduct a statistical analysis on all sources.
From each unwhitened and whitened residual of each source, we collect 100 patches of size $16\times 16$. 
From these patches, we compute the mean $\mu_i$ and covariance $\Sigma_i$ for each source $i$. 
We assign a Gaussian distribution $\mathcal{N}(\mu_i,\Sigma_i)$ to each source $i$ and compute the pairwise Jeffreys divergences between all sources.
Finally, we feed the resulting dissimilarity matrix to the multidimensional scaling (MDS) algorithm~\cite{cox2008multidimensional} to embed each distribution into 2D coordinates.
Fig. 4 shows this embedding.
Before whitening, most sources have arbitrary distances with respect to each other. 
Whitening strongly reduces OOD bias, juxtaposing the generated sources while separating the real source \texttt{raise1k}, which explains the performance boost.  
In contrast, \texttt{firefly} and \texttt{glide} become clear outliers after whitening, leading to inferior generalization performance on these datasets.
They are already known for their distinct 2D Fourier representations, where \texttt{glide} has few periodic patterns, while \texttt{firefly} has excessive periodic peaks~\cite{bammey2023synthbuster}.
The analysis in Fig.~\ref{fig:mds-embeddings} implies that certain artifacts in \texttt{firefly} and \texttt{glide} remain prominent even after whitening. 
The reason why the whitening cannot eliminate these artifacts can be investigated by further analysis on the patches\textemdash by going beyond the scope of this work.

\section{Conclusion}
\label{sec:conclusion}
We propose using a deconvolution process to whiten image noise residuals, improving the generalisation power of generated image detectors.
The deconvolution process estimates and removes local correlations in noise residuals, which often contain specific generator fingerprints to which AI detectors overfit, reducing their generalization capacity.
While this work provides a proof of concept for the effectiveness of residual whitening, it opens several avenues worth exploring. 
One of these is the choice of the residual extractor. 
Because the proposed approach is compatible with any type of residual extractor, it is worth investigating the effect of whitening on other, more sophisticated\textemdash deep model residual extractors.
Another direction is to go beyond generalization\textemdash this analysis can be useful for image attribution (\emph{i.e.} multi-class classification of generators) or anti-forensic tasks (\emph{e.g.} adversarial attacks on AI detectors), which we plan to develop further in the future. 

\clearpage
\section{Acknowledgements}
This work was supported by the Agence National de la Recherche through the CI2(IA) project ANR-23-IAS4-0004 and the France 2030 program ANR-22-PECY-0011.

\bibliographystyle{IEEEbib}
\bibliography{strings,refs}

\begin{thebibliography}{10}

\bibitem{shoaib2023deepfakes}
Mohamed~R Shoaib, Zefan Wang, Milad~Taleby Ahvanooey, and Jun Zhao,
\newblock ``Deepfakes, misinformation, and disinformation in the era of
  frontier {AI}, generative {AI}, and large {AI} models,''
\newblock in {\em 2023 international conference on computer and applications
  (ICCA)}. IEEE, 2023, pp. 1--7.

\bibitem{cozzolino2024raising}
Davide Cozzolino, Giovanni Poggi, Riccardo Corvi, Matthias Nie{\ss}ner, and
  Luisa Verdoliva,
\newblock ``Raising the bar of {AI}-generated image detection with {CLIP},''
\newblock in {\em 2024 IEEE/CVF Conference on Computer Vision and Pattern
  Recognition Workshops (CVPRW)}. IEEE, 2024, pp. 4356--4366.

\bibitem{baraldi2024contrasting}
Lorenzo Baraldi, Federico Cocchi, Marcella Cornia, Lorenzo Baraldi, Alessandro
  Nicolosi, and Rita Cucchiara,
\newblock ``Contrasting deepfakes diffusion via contrastive learning and
  global-local similarities,''
\newblock in {\em European Conference on Computer Vision}. Springer, 2024, pp.
  199--216.

\bibitem{guillaro2025bias}
Fabrizio Guillaro, Giada Zingarini, Ben Usman, Avneesh Sud, Davide Cozzolino,
  and Luisa Verdoliva,
\newblock ``A bias-free training paradigm for more general {AI}-generated image
  detection,''
\newblock in {\em 2025 IEEE/CVF Conference on Computer Vision and Pattern
  Recognition (CVPR)}. IEEE, 2025, pp. 18685--18694.

\bibitem{bammey2023synthbuster}
Quentin Bammey,
\newblock ``Synthbuster: Towards detection of diffusion model generated
  images,''
\newblock {\em IEEE Open Journal of Signal Processing}, vol. 5, pp. 1--9, 2023.

\bibitem{mallet2025simple}
Antoine Mallet, Arthur M{\'e}reur, Minoru Kuribayashi, R{\'e}mi Cogranne, and
  Patrick Bas,
\newblock ``Simple detection of {AI}-generated images based on noise
  correlation,''
\newblock in {\em 2025 International Conference on Advanced Machine Learning
  and Data Science (AMLDS)}. IEEE, 2025, pp. 265--270.

\bibitem{francu2026cover}
Jan Francu, Milan Papez, and Tomas Pevny,
\newblock ``Cover-source mismatch in deepfake detection: A systematic study,''
\newblock in {\em Proceedings of the 2026 ACM Workshop on Information Hiding
  and Multimedia Security}, 2026, pp. 28--39.

\bibitem{uhlenbrock2024did}
Lea Uhlenbrock, Davide Cozzolino, Denise Moussa, Luisa Verdoliva, and Christian
  Riess,
\newblock ``Did you note my palette? unveiling synthetic images through color
  statistics,''
\newblock in {\em Proceedings of the 2024 ACM Workshop on Information Hiding
  and Multimedia Security}, 2024, pp. 47--52.

\bibitem{li2024masksim}
Yanhao Li, Quentin Bammey, Marina Gardella, Tina Nikoukhah, Jean-Michel Morel,
  Miguel Colom, and Rafael~Grompone Von~Gioi,
\newblock ``Masksim: Detection of synthetic images by masked spectrum
  similarity analysis,''
\newblock in {\em 2024 IEEE/CVF Conference on Computer Vision and Pattern
  Recognition Workshops (CVPRW)}. IEEE, 2024, pp. 3855--3865.

\bibitem{ye2019network}
Chengxi Ye, Matthew Evanusa, Hua He, Anton Mitrokhin, Tom Goldstein, James~A
  Yorke, Cornelia Ferm{\"u}ller, and Yiannis Aloimonos,
\newblock ``Network deconvolution,''
\newblock {\em arXiv preprint arXiv:1905.11926}, 2019.

\bibitem{li2025detecting}
Ji~Li and Kai Wang,
\newblock ``{Detecting computer-generated images by using only real images},''
\newblock in {\em International Conference on Machine Vision}, Wolfgang Osten,
  Ed. International Society for Optics and Photonics, 2025, vol. 13517, p.
  135170Q, SPIE.

\bibitem{abecidan2022using}
Rony Abecidan, Vincent Itier, J{\'e}r{\'e}mie Boulanger, Patrick Bas, and
  Tom{\'a}{\v{s}} Pevn{\'y},
\newblock ``Using set covering to generate databases for holistic
  steganalysis,''
\newblock in {\em 2022 IEEE International Workshop on Information Forensics and
  Security (WIFS)}. IEEE, 2022, pp. 1--6.

\bibitem{abecidan2026tackle}
Rony Abecidan, Vincent Itier, J{\'e}r{\'e}mie Boulanger, Patrick Bas, and
  Tom{\'a}{\v{s}} Pevn{\`y},
\newblock ``Tackle {CSM} in {JPEG} steganalysis with data adaptation,''
\newblock in {\em Proceedings of the 2026 ACM Workshop on Information Hiding
  and Multimedia Security}, 2026, pp. 121--126.

\bibitem{corvi2023intriguing}
Riccardo Corvi, Davide Cozzolino, Giovanni Poggi, Koki Nagano, and Luisa
  Verdoliva,
\newblock ``{ Intriguing properties of synthetic images: from generative
  adversarial networks to diffusion models },''
\newblock in {\em IEEE/CVF Conference on Computer Vision and Pattern
  Recognition Workshops (CVPRW)}. 2023, pp. 973--982, IEEE Computer Society.

\bibitem{elad2023image}
Michael Elad, Bahjat Kawar, and Gregory Vaksman,
\newblock ``Image denoising: The deep learning revolution and beyond—a survey
  paper,''
\newblock {\em SIAM Journal on Imaging Sciences}, vol. 16, no. 3, pp.
  1594--1654, 2023.

\bibitem{kessy2018optimal}
Agnan Kessy, Alex Lewin, and Korbinian Strimmer,
\newblock ``Optimal whitening and decorrelation,''
\newblock {\em The American Statistician}, vol. 72, no. 4, pp. 309--314, 2018.

\bibitem{huang2018decorrelated}
Lei Huang, Dawei Yang, Bo~Lang, and Jia Deng,
\newblock ``Decorrelated batch normalization,''
\newblock in {\em Proceedings of the IEEE conference on computer vision and
  pattern recognition}, 2018, pp. 791--800.

\bibitem{doi2025deepfake}
Minh~Thong Doi, Jan Butora, Vincent Itier, J{\'e}r{\'e}mie Boulanger, and
  Patrick Bas,
\newblock ``Deepfake detection based on noise residuals,''
\newblock in {\em GRETSI’25}, 2025.

\bibitem{JIN21}
Jan Butora, Yassine Yousfi, and Jessica Fridrich,
\newblock ``How to pretrain for steganalysis,''
\newblock in {\em Proceedings of the 2021 ACM Workshop on Information Hiding
  and Multimedia Security}, 2021, p. 143–148.

\bibitem{dang2015raise}
Duc-Tien Dang-Nguyen, Cecilia Pasquini, Valentina Conotter, and Giulia Boato,
\newblock ``Raise: A raw images dataset for digital image forensics,''
\newblock in {\em Proceedings of the 6th ACM multimedia systems conference},
  2015, pp. 219--224.

\bibitem{bled2023pushing}
Cl{\'e}ment Bled and Fran{\c{c}}ois Piti{\'e},
\newblock ``Pushing the limits of the wiener filter in image denoising,''
\newblock in {\em 2023 IEEE International conference on image processing
  (ICIP)}. IEEE, 2023, pp. 2590--2594.

\bibitem{cox2008multidimensional}
Michael~AA Cox and Trevor~F Cox,
\newblock ``Multidimensional scaling,''
\newblock in {\em Handbook of data visualization}, pp. 315--347. Springer,
  2008.

\end{thebibliography}

\end{document}